\documentclass[10pt]{article} 
\usepackage[preprint]{tmlr}

\usepackage{amsmath,amsfonts,bm}

\def\eqref#1{equation~\ref{#1}}

\def\1{\bm{1}}

\DeclareMathAlphabet{\mathsfit}{\encodingdefault}{\sfdefault}{m}{sl}
\SetMathAlphabet{\mathsfit}{bold}{\encodingdefault}{\sfdefault}{bx}{n}

\usepackage{hyperref}
\usepackage{url}
\usepackage{booktabs}
\usepackage{multirow}
\usepackage{graphicx}
\usepackage{enumitem}
\usepackage{subcaption}
\usepackage{xcolor}
\definecolor{gainGreen}{HTML}{1E8449}
\usepackage{amsmath}
\usepackage{amssymb}
\usepackage{algorithm}
\usepackage{algpseudocode}
\usepackage{colortbl}
\usepackage{placeins}
\usepackage{float}
\usepackage{flafter}

\title{CanvasAnneal: Curriculum Reinforcement Learning for \\ Diffusion Language Models}

\author{\name Blake Olson \email blakeolson@utexas.edu \\
      \addr Google DeepMind
      \AND
      \name Yuhang Song \email yuhangsong@google.com \\
      \addr Google DeepMind
      \AND
      \name Emmett McQuinn \email emmettmcquinn@google.com \\
      \addr Google DeepMind
      \AND
      \name Yuan Shangguan \email yuansg@google.com \\
      \addr Google DeepMind \\
      }

\def\month{MM}  
\def\year{YYYY} 
\def\openreview{\url{https://openreview.net/forum?id=XXXX}} 

\begin{document}

\maketitle

\begin{abstract}
Diffusion Language Models (DLMs) offer promising parallel generation capabilities but lag behind autoregressive models in complex reasoning and tool-use tasks. While Reinforcement Learning (RL) has recently been applied to enhance DLMs, standard RL approaches suffer from an exploration bottleneck. To address this, we inject reasoning priors from a stronger teacher model to guide RL exploration. In this paper, we introduce CanvasAnneal, a curriculum-guided diffusion RL framework. During the initial RL phase, we warm-start exploration by injecting teacher-generated reasoning traces into the initial diffusion canvas. As training progresses, we gradually remove this guidance and require the model to generate more of the reasoning trajectory independently. Across mathematical reasoning and tool-use benchmarks, CanvasAnneal improves over standard diffu-GRPO on MATH500, Countdown, and Tau2 and substantially accelerates reward improvement on several tasks, while gains are task-dependent. Our results suggest that structured training-time guidance can alleviate exploration bottlenecks in diffusion RL and speed up convergence on harder tasks.
\end{abstract}

\section{Introduction}

Diffusion Language Models (DLMs) have emerged as a compelling alternative to classical Autoregressive Models (ARMs) for text generation, structured sequence modeling \citep{austin2023structureddenoisingdiffusionmodels, lou2024discretediffusionmodelingestimating} and complex reasoning tasks \citep{zhao2025d1scalingreasoningdiffusion, zhou2026dllmsimplediffusionlanguage}. ARMs generate text token-by-token in a rigid left-to-right causal order. In contrast, DLMs, such as LLaDA \citep{nie2025largelanguagediffusionmodels} and Dream \citep{ye2025dream7bdiffusionlarge}, employ discrete or masked diffusion processes to refine complete sequences in parallel through iterative denoising. By removing unidirectional constraints, DLMs offer distinct theoretical and practical advantages. These advantages enable arbitrary-order infilling, parallel iterative editing, flexible speed-quality trade-offs, and relief from sequential alignment issues like the reversal curse \citep{nie2025largelanguagediffusionmodels}.

Although DLMs possess structural strengths, scaling them to match state-of-the-art ARMs on complex multi-step reasoning and tool-use tasks remains a significant challenge \citep{lu2026bitterlessondiffusionlanguage}. Post-training paradigms, including Supervised Fine-Tuning (SFT) and Reinforcement Learning (RL), have driven substantial reasoning gains in ARMs. However, extending RL to discrete diffusion architectures presents practical bottlenecks. In ARMs, the policy executes step-by-step token selection with clear causal credit assignment. In contrast, DLMs start sampling from a fully masked, unconditioned noise canvas $\mathbf{x}_T$, which requires credit assignment across both spatial token positions and temporal reverse diffusion steps. High entropy in noise space causes significant variance in trajectory rollouts during early RL iterations, preventing the policy from discovering viable reasoning traces or stable tool call executions \citep{zhao2025d1scalingreasoningdiffusion}.

To bridge this performance gap, recent works have explored various algorithmic and objective approximations. Policy and reward approximations optimize step-wise likelihood lower bounds or marginal probabilities, enabling reward signal propagation without calculating exact marginal likelihoods over all reverse trajectories \citep{zhao2025d1scalingreasoningdiffusion}. Trajectory and distillation approaches improve diffusion generation by simplifying reverse trajectories or transferring knowledge from autoregressive models, improving sampling efficiency and training stability \citep{zhang2026fewstepdiffusionlanguagemodels, qian2026d3llmultrafastdiffusionllm, su2026dataefficientautoregressivetodiffusionlanguagemodels}. Meanwhile, inference-time approaches incorporate formal grammar constraints and lookahead verification to steer noisy sampling toward syntactically valid outputs \citep{suresh2025dingoconstrainedinferencediffusion, zhang2026lookaheadthenverifyreliableconstraineddecoding}. Selective supervision frameworks apply adaptive loss masking, which focuses update gradients primarily on high-value or low-entropy sequence regions \citep{chen2026trimstrajectoryrankedinstructionmasked, parashar2026learnabilityinformedfinetuningdiffusionlanguage}.

Although these approximations improve overall stability, standard RL exploration still suffers when initialized entirely from pure noise without structural guidance. This makes intuitive sense, since noise sampling in early stages lacks directional feedback \citep{engels2026transparentdiffusiongemma}. Motivated by this exploration bottleneck, we develop CanvasAnneal. We use structured reasoning priors generated by a stronger teacher model and inject them into the initial diffusion canvas, thereby warm-starting the diffusion process during early RL training. As training progresses, this structural scaffolding is annealed through a dynamic stochastic curriculum, progressively reducing the amount of teacher-provided context until generation is initialized from an almost fully masked canvas.

Our contributions are threefold. First, we introduce a curriculum for diffusion RL that varies the amount of teacher-provided reasoning context through suffix masking while retaining fully unassisted rollouts throughout training. Second, we integrate this curriculum with group-relative policy optimization by sharing the same initialization across all rollouts within a prompt group. Third, we evaluate the method on mathematical reasoning and tool-use tasks. CanvasAnneal improves over standard diffu-GRPO on MATH500 and Countdown across the evaluated generation lengths and accelerates reward improvement on xLAM and Countdown. Results on GSM8K and tool-use benchmarks are mixed, indicating that the benefit of curriculum-guided initialization is task-dependent.

\section{Related Work}

\textbf{Diffusion Language Models.} 
Diffusion Language Models (DLMs) have emerged as scalable non-autoregressive architectures capable of parallel sequence refinement \citep{austin2023structureddenoisingdiffusionmodels, lou2024discretediffusionmodelingestimating, sahoo2024simpleeffectivemaskeddiffusion}. Modern discrete diffusion models, such as LLaDA and Dream, corrupt categorical tokens directly via masked transitions, enabling dynamic infilling and flexible speed-quality trade-offs \citep{nie2025largelanguagediffusionmodels, ye2025dream7bdiffusionlarge}. However, training DLMs for complex reasoning tasks remains challenging due to the intractable nature of exact sequence likelihoods.

\textbf{Reinforcement Learning for DLMs.}
Adapting Reinforcement Learning (RL) to discrete diffusion is challenging because credit assignment spans iterative denoising steps and multiple token positions
\citep{zhao2025d1scalingreasoningdiffusion, huang2025reinforcingdiffusionchainlateral, he2025mdpoovercomingtraininginferencedivide}.
Group Relative Policy Optimization (GRPO) avoids the need for a separate critic by estimating advantages from the relative rewards of multiple sampled responses to the same prompt
\citep{shao2024deepseekmathpushinglimitsmathematical}.
Diffu-GRPO adapts this framework to diffusion language models using tractable masked likelihood approximations
\citep{zhao2025d1scalingreasoningdiffusion},
while DCOLT and MDPO optimize intermediate diffusion states or trajectories
\citep{huang2025reinforcingdiffusionchainlateral, he2025mdpoovercomingtraininginferencedivide}.
More recent approaches seek more faithful or stable policy optimization through trajectory-likelihood estimation
\citep{wang2026d2improvingreasoningdiffusion},
explicit modeling of masking decisions
\citep{raajesh2026maskawarepolicygradientsdiffusion},
multi-step diffusion policy-gradient estimation
\citep{zhan2026simplepolicygradientsreasoning},
and improved clipping and normalization for noisy likelihood-ratio estimates
\citep{zhong2026stabilizingreinforcementlearningdiffusion}.
Other methods reformulate the optimization objective more substantially: wd1 replaces importance-ratio optimization with a weighted log-likelihood objective
\citep{tang2026wd1weightedpolicyoptimization},
while GDSD casts reinforcement learning as guided denoiser self-distillation
\citep{tang2026gdsdreinforcementlearningguided}.

\textbf{Teacher-guided RL exploration.}
Recent methods such as IGPO \citep{zhao2025inpaintingguidedpolicyoptimizationdiffusion} and BREAD \citep{zhang2025breadbranchedrolloutsexpert} have demonstrated that injecting partial reasoning traces can guide RL exploration and prevent sample waste. However, these methods differ structurally from CanvasAnneal across several axes. IGPO introduces teacher information into diffusion models via scattered inpainting masks adaptively based on policy failure. BREAD applies causal reasoning prefixes to standard autoregressive models, adaptively branching rollouts from expert anchors. In contrast, CanvasAnneal integrates teacher information into the initial diffusion canvas using contiguous suffix masking, and it fades guidance globally using a training-progress-dependent Beta curriculum rather than adapting to per-rollout failures. Furthermore, CanvasAnneal explicitly retains a fixed fraction of pure-noise rollouts ($p_{\text{pure}}$) throughout training. By applying the exact same partially masked canvas across all $G$ completions, CanvasAnneal ensures a strictly controlled group-relative advantage computation. Like IGPO and BREAD, our method relies on the teacher only during training and requires no auxiliary models at inference.

\textbf{Curriculum Learning.} 
Curriculum learning smooths optimization by progressively introducing complex training conditions \citep{10.1145/1553374.1553380} \citep{bengio2015scheduledsamplingsequenceprediction}. Recently, task-level curricula have been applied to RL post-training to schedule prompt difficulties from easy to hard \citep{parashar2026curriculumreinforcementlearningeasy}. In non-autoregressive regimes, however, models require structural guidance over sequence corruption rather than dataset-level filtering. Building on teacher-trace annealing \citep{bengio2015scheduledsamplingsequenceprediction}, we formulate a dynamic masking curriculum over the diffusion canvas, progressively closing the gap between teacher-guided scaffolding and autonomous denoising without requiring task-difficulty annotations or supervised fine-tuning pipelines.

\section{Preliminaries}

\paragraph{Discrete Masked Diffusion Language Models.}
Let $\mathbf{x}_0 = (x_{0,1}, \dots, x_{0,N}) \in \mathcal{V}^N$ denote a target text sequence of length $N$ over a vocabulary $\mathcal{V}$. Discrete masked diffusion models define a forward process $q(\mathbf{x}_t \mid \mathbf{x}_0)$ over discrete timesteps $t \in \{1, \dots, T\}$ by corrupting tokens into a designated $\text{[MASK]}$ token:
\begin{equation}
\begin{aligned}
q(\mathbf{x}_t \mid \mathbf{x}_0) &= \prod_{j=1}^N q(x_{t,j} \mid x_{0,j}), \\
q(x_{t,j} \mid x_{0,j}) &= (1 - \gamma_t)\delta_{x_{0,j}} + \gamma_t \delta_{\text{[MASK]}}
\end{aligned}
\end{equation}
where $\gamma_t \in [0, 1]$ is a monotonically increasing masking schedule such that $\mathbf{x}_T$ consists entirely of $\text{[MASK]}$ tokens, and $\delta$ denotes the Kronecker delta (point mass). 

The generative reverse process is parameterized by a neural network
$p_\theta(\mathbf{x}_0 \mid \mathbf{x}_t)$, which predicts clean sequence
logits from a corrupted state $\mathbf{x}_t$. For simplicity, we describe
DLM training using the following masked-token cross-entropy objective:
\begin{equation}
\mathcal{L}_{\text{DLM}}(\theta) =
\mathbb{E}_{t, \mathbf{x}_0, \mathbf{x}_t}
\left[
\sum_{j \in \mathcal{M}(\mathbf{x}_t)}
-\log p_\theta(x_{0,j} \mid \mathbf{x}_t)
\right],
\end{equation}
where $\mathcal{M}(\mathbf{x}_t) = \{j: x_{t,j} = \text{[MASK]}\}$
denotes the indices of masked tokens.

\paragraph{Group Relative Policy Optimization for DLMs.}
In Group Relative Policy Optimization (GRPO), a prompt $\mathbf{q}$ generates a group of
$G$ candidate responses $\{y_1, y_2, \dots, y_G\}$.
Because exact sequence log-likelihoods $\log \pi_\theta(y_i \mid \mathbf{q})$
are intractable for non-autoregressive diffusion models, diffu-GRPO uses a
one-step approximation to the per-token likelihood. Specifically, a perturbed
prompt $\mathbf{q}'$ is constructed by randomly masking prompt tokens, while the
entire response is replaced with $\text{[MASK]}$ tokens. For a response
$y_i = (y_{i,1}, \dots, y_{i,N_i})$, the approximate token log-likelihood is
\begin{equation}
\hat{\ell}_\theta(y_{i,j} \mid \mathbf{q}')
=
\log p_\theta
\left(
y_{i,j}
\mid
\mathbf{q}' \oplus \text{[MASK]}^{N_i}
\right),
\qquad j = 1,\dots,N_i,
\end{equation}
where $\oplus$ denotes sequence concatenation.

Each response $y_i$ receives a scalar reward
$R_i = R(y_i, \mathbf{q})$.
Following diffu-GRPO, we compute the unnormalized group-relative advantage as
\begin{equation}
A_i =
R_i - \frac{1}{G}\sum_{g=1}^{G} R_g.
\end{equation}
The same sequence-level advantage $A_i$ is assigned to every token in response $y_i$.

Instead of an exact sequence-level importance-sampling ratio, diffu-GRPO computes a token-level importance ratio using the one-step likelihood approximation under the current policy $\theta$ and the old policy $\theta_{\mathrm{old}}$:
\begin{equation}
r_{i,j}(\theta)
=
\exp \left(
\hat{\ell}_\theta(y_{i,j} \mid \mathbf{q}')
-
\hat{\ell}_{\theta_{\mathrm{old}}}(y_{i,j} \mid \mathbf{q}')
\right),
\end{equation}
Using this approximation, diffu-GRPO applies PPO-style clipping \citep{schulman2017proximalpolicyoptimizationalgorithms} to the
token-level importance ratios together with a KL penalty toward a fixed
reference policy. The resulting objective is
\begin{equation}
\begin{aligned}
\mathcal{J}_{\text{diffu-GRPO}}(\theta)
= \mathbb{E}\Bigg[
&\frac{1}{G}\sum_{i=1}^{G}\frac{1}{N_i}\sum_{j=1}^{N_i}
\min\!\left(r_{i,j}(\theta)A_i,\,
\operatorname{clip}(r_{i,j}(\theta),1-\epsilon,1+\epsilon)A_i\right) \\
&-\beta D_{\mathrm{KL}}\!\left[
\phi_\theta(\cdot\mid\mathbf{q}')
\,\|\,\phi_{\mathrm{ref}}(\cdot\mid\mathbf{q}')
\right]
\Bigg].
\end{aligned}
\end{equation}

\section{Methods}

\begin{figure}[t]
\centering
\includegraphics[width=\linewidth]{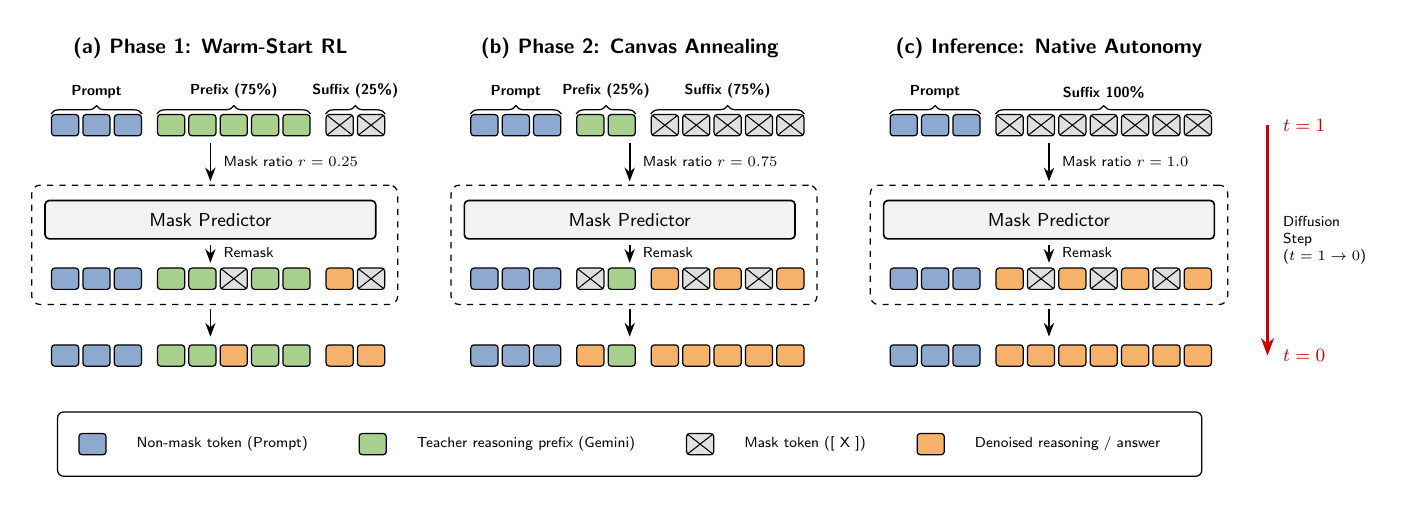}
\caption{\textbf{Overview of CanvasAnneal via canvas annealing.} (a) \textbf{Warm-start RL.} Training can begin from a suffix-masked teacher reasoning trace. For example, with $m=0.25$, the model observes a reasoning prefix while generating the remaining suffix. (b) \textbf{Canvas annealing.} As training progresses, the masking ratio increases, reducing the amount of teacher-provided context and requiring the policy to complete larger portions of the reasoning trajectory. (c) \textbf{Inference.} At test time, the teacher is removed and generation begins from a fully masked canvas ($m=1.0$), preserving native diffusion inference.}
\label{fig:overview}
\end{figure}

Existing diffusion RL frameworks provide tractable surrogate objectives, but optimizing policies from fully masked canvases $\mathbf{x}_T$ can enhance exploration difficulty and sparse reward signals, particularly early in training \citep{zhao2025inpaintingguidedpolicyoptimizationdiffusion, zhang2025breadbranchedrolloutsexpert}. Alternatively, providing teacher rationales during inference can improve reasoning performance but introduces additional latency and removes part of the parallel-efficiency advantage of diffusion models. We introduce CanvasAnneal, which uses teacher reasoning traces as training-time scaffolds and progressively removes this guidance over the course of optimization. Our method consists of four components: teacher reasoning trace generation, reasoning injection into the diffusion canvas, a dynamic masking curriculum, and group-level policy optimization. Figure~\ref{fig:overview} provides an overview of the training procedure.

\subsection{Teacher Reasoning Trace Generation}

We generate offline reference reasoning traces using Gemini 3.1 Pro. Given a problem prompt $\mathbf{q}$, the teacher produces a structured rationale $\boldsymbol{\tau} = (\tau_1, \tau_2, \dots, \tau_M) \in \mathcal{V}^M$.

We construct dataset-specific instruction templates that request intermediate derivation steps while withholding the final numerical answer or tool execution command. Teacher outputs are length-calibrated to approximately 200 words, or roughly 256 tokens, to reduce truncation when the trace is placed into the diffusion canvas. Mathematical rationales are enclosed within $\texttt{<reasoning>} \dots \newline \texttt{</reasoning>}$ tags, whereas function-calling rationales use $\texttt{<think>} \dots \texttt{</think>}$ tags. We prompt the model to not include any final answer within the reasoning.

\FloatBarrier

\subsection{Reasoning Injection into Diffusion Canvas}
Standard diffusion sampling initializes generation from a sequence of mask tokens,
$\mathbf{x}_T = (\text{[MASK]})^N$.
Our approach instead uses a teacher reasoning trace $\boldsymbol{\tau}$ to partially initialize the completion canvas during training.

Let $N$ denote the maximum completion length, and let $M' = \min(M, N)$ denote the allocated trace length. We initialize completion canvas $\mathbf{x}_{\text{init}} \in \mathcal{V}^N$ by placing teacher tokens into the first $M'$ positions:
\begin{equation}
x_{\text{init}, j} = 
\begin{cases} 
\tau_j, & \text{for } 1 \le j \le M', \\
\text{[MASK]}, & \text{for } M' < j \le N.
\end{cases}
\end{equation}

A masking fraction $m \in [0,1]$ determines how much of the injected reasoning trace is re-masked. Independent random masking can remove tokens throughout the rationale, producing fragmented intermediate states. We instead use suffix masking, which preserves a contiguous prefix of the teacher trajectory.

For each position $j \in \{1,\dots,M'\}$, we define the rank score
\begin{equation}
s_j = 1 - \frac{j}{M'}.
\end{equation}
A token at position $j$ is replaced with $\text{[MASK]}$ when $s_j < m$, equivalently when
$j > \lfloor(1-m)M'\rfloor$.
The resulting canvas therefore preserves the prefix
$(\tau_1,\dots,\tau_{\lfloor(1-m)M'\rfloor})$
and masks the remaining suffix.

This construction exposes the model to a coherent prefix of the reference reasoning trajectory while requiring it to generate the remaining reasoning steps and final answer. As $m$ increases, the amount of teacher-provided context decreases.

During CanvasAnneal generation, the model conditions on a set of fixed teacher positions $C_g$ to produce the generated positions $U_g$. However, when computing the GRPO policy objective, the surrogate likelihood evaluates over the full completion sequence containing both the teacher prefix and the generated tokens. Let $y_i$ represent the concatenation of positions in $C_g \cup U_g$. The conditioned-surrogate objective evaluates all positions, meaning both $C_g$ and $U_g$ are subject to masking and enter the cross-entropy loss:

\begin{equation}
\hat{\ell}_\theta(y_{i,j} \mid \mathbf{q}')
=
\log p_\theta
\left(
y_{i,j}
\mid
\mathbf{q}' \oplus \text{[MASK]}^{N_i}
\right) \quad \text{for } j \in C_g \cup U_g,
\end{equation}

where $y_{i,j}$ includes both the generated text and the fixed teacher prefix.

While including the fixed teacher prefix $C_g$ alongside the generated tokens $U_g$ in the surrogate objective (Equation 9) diverges from standard autoregressive policy gradients, this design is grounded in the parallel nature of discrete diffusion. In diffusion language models, credit assignment spans across both temporal reverse diffusion steps and spatial token positions simultaneously. Structurally separating the prefix from the suffix during the cross-entropy evaluation is unnatural for parallel iterative denoising. Importantly, this inclusion preserves the controlled comparison underlying the group-relative update. Because the sampled mask ratio $m_g$ and the resulting initial canvas $x_{init}^{(g)}$ are shared exactly across all $G$ rollouts in a prompt group, the group-relative advantage computation remains perfectly isolated. Therefore, any variance in the advantage is driven exclusively by the policy's autonomous generation in $U_g$. Ultimately, evaluating the full sequence $C_g \cup U_g$ acts as an advantage-weighted distillation of the teacher's rationale \citep{su2026dataefficientautoregressivetodiffusionlanguagemodels, tang2026gdsdreinforcementlearningguided}, reinforcing the model's ability to reconstruct the expert prior when it leads to a highly rewarded trajectory.

\subsection{Curriculum Annealing via Dynamic Beta Mixture}
To ensure the policy transitions from teacher-guided exploration to autonomous generation, we formulate a dynamic curriculum schedule over mask fraction $m$.

\paragraph{Mixture Distribution.}
At global training step $k \in \{1, \dots, K\}$, we define training progress as $\rho_k = \frac{k}{K} \in [0, 1]$. To prevent the policy from developing an exclusive reliance on teacher prefixes, we construct a mixture distribution at the prompt-group level. For each prompt group, a Bernoulli trial with probability $p_{\text{pure}}$ assigns a pure noise canvas ($m = 1.0$). With probability $1 - p_{\text{pure}}$, the mask ratio $m$ is sampled from a time-varying Beta distribution:
\begin{equation}
\mathcal{D}_{\text{curriculum}}(k) = p_{\text{pure}} \, \delta_1(m) + (1 - p_{\text{pure}}) \, \text{Beta}\left( m; \, \alpha_k, \beta_k \right)
\end{equation}
where $\delta_1$ denotes a Dirac point mass at $m = 1.0$. For CanvasAnneal, we use $p_{\text{pure}}=0.25$ for xLAM and $p_{\text{pure}}=0.75$ for all math reasoning benchmarks.

\paragraph{Dynamic Beta Evolution.}
The Beta distribution parameters $\alpha_k$ and $\beta_k$ evolve linearly with training progress $\rho_k$:
\begin{equation}
\alpha_k = \alpha_0 + (\alpha_{\text{end}} - \alpha_0) \rho_k, \quad \beta_k = \beta_0 + (\beta_{\text{end}} - \beta_0) \rho_k
\end{equation}
We initialize parameters with $\alpha_0 = 1.0$, $\alpha_{\text{end}} = 5.0$, $\beta_0 = 5.0$, and $\beta_{\text{end}} = 1.0$. During initial training iterations ($\rho_k \approx 0$), $\text{Beta}(1, 5)$ has expected mask fraction $\mathbb{E}[m]=1/6\approx 0.167$, retaining approximately $83\%$ of the teacher trace and providing dense reward feedback. Near training convergence ($\rho_k \to 1$), $\text{Beta}(5, 1)$ has expected mask fraction $\mathbb{E}[m]=5/6\approx 0.833$, requiring the policy to generate a larger fraction of the reasoning trajectory independently.

\paragraph{Group-Level Canvas Consistency.}
In Group Relative Policy Optimization, each prompt $\mathbf{q}$ generates a group of $G$ candidate completions $\{y_1, \dots, y_G\}$. In our framework, the sampled mask ratio $m_g$ and the resulting initial canvas $\mathbf{x}_{\text{init}}^{(g)}$ are shared across all $G$ rollouts in group $g$. This ensures all candidate rollouts are evaluated under identical context constraints, resulting in fair intra-group advantage computation.

\subsection{Policy Optimization and Native Inference}
Given prompt $\mathbf{q}$ and initial canvas $\mathbf{x}_{\text{init}}$, the policy network $p_\theta$ performs $S$ iterative denoising steps to generate rollouts $y_i$. Completions are scored by task-specific reward functions $R(y_i, \mathbf{q})$, and advantages $A_i$ are computed across the group. The complete training procedure is summarized in Algorithm~\ref{alg:curriculum_diffu_grpo}.

At test time, the teacher model is entirely omitted. The model samples candidate outputs directly from an unconditioned noise canvas ($m = 1.0$) using native iterative denoising, without additional teacher inference.

\begin{algorithm}[t]
\caption{CanvasAnneal Training}
\label{alg:curriculum_diffu_grpo}
\begin{algorithmic}[1]
\Require Dataset $\mathcal{D} = \{(\mathbf{q}, \boldsymbol{\tau}, a)\}$, base policy $\pi_\theta$, total steps $K$, group size $G$, mixture probability $p_{\text{pure}}$, Beta parameters $(\alpha_0, \alpha_{\text{end}}, \beta_0, \beta_{\text{end}})$.
\For{step $k = 1$ \textbf{to} $K$}
    \State Sample mini-batch of prompts, teacher traces, and answers $\{(\mathbf{q}_g, \boldsymbol{\tau}_g, a_g)\}_{g=1}^B \sim \mathcal{D}$
    \State Compute training progress $\rho_k \leftarrow k / K$
    \State Update Beta parameters $\alpha_k \leftarrow \alpha_0 + (\alpha_{\text{end}} - \alpha_0)\rho_k$, $\beta_k \leftarrow \beta_0 + (\beta_{\text{end}} - \beta_0)\rho_k$
    \For{each group $g \in \{1, \dots, B\}$}
        \State Sample $u \sim \mathcal{U}(0, 1)$
        \If{$u < p_{\text{pure}}$}
            \State Set mask ratio $m_g \leftarrow 1.0$
        \Else
            \State Sample mask ratio $m_g \sim \text{Beta}(\alpha_k, \beta_k)$
        \EndIf
        \State Construct suffix-masked initial canvas $\mathbf{x}_{\text{init}}^{(g)}$ via Eq.~(7) and Eq.~(8)
        \State Generate $G$ parallel rollouts $\{y_{g,1}, \dots, y_{g,G}\}$ using $\pi_\theta(\cdot \mid \mathbf{q}_g, \mathbf{x}_{\text{init}}^{(g)})$
        \State Evaluate rewards $R_{g,i} \leftarrow R(y_{g,i}, \mathbf{q}_g, a_g)$ for $i \in \{1, \dots, G\}$
        \State Compute group advantages
        $A_{g,i} \leftarrow R_{g,i} - \frac{1}{G}\sum_{j=1}^{G} R_{g,j}$
    \EndFor
    \State Construct perturbed prompt $\mathbf{q}'_g$ by randomly masking prompt tokens
    \State Replace the completion with $\text{[MASK]}$ tokens
    \State Evaluate one-step token likelihoods under current $\pi_\theta$ and old $\pi_{\theta_{\mathrm{old}}}$ policies
    \State Compute token-level surrogate ratios
    $r_{g,i,j}(\theta) \leftarrow
    \exp\left(
    \hat{\ell}_\theta(y_{g,i,j} \mid \mathbf{q}'_g)
    -
    \hat{\ell}_{\theta_{\mathrm{old}}}(y_{g,i,j} \mid \mathbf{q}'_g)
    \right)$
    \State Update policy parameters $\theta$ by maximizing objective
    $\mathcal{J}_{\text{diffu-GRPO}}(\theta)$ via Eq.~(6)
\EndFor
\end{algorithmic}
\end{algorithm}

\section{Experiments}

We evaluate CanvasAnneal on mathematical reasoning and tool-use tasks. Our experiments examine three questions: whether curriculum-guided initialization improves downstream task accuracy, whether it affects RL optimization dynamics, and whether teacher-guided RL can provide benefits without requiring teacher inference at test time.

\subsection{Experimental Setup}

\paragraph{Backbone Model.}
We employ LLaDA-7B-A1B-Instruct \citep{zhu2025lladamoesparsemoediffusion} as our primary base architecture. LLaDA is an open-weights discrete diffusion language model featuring 7B parameters, designed for parallel text generation and arbitrary infilling via masked diffusion.

\paragraph{Datasets and Tasks.}
We evaluate model performance across diverse mathematical reasoning and tool-use benchmarks:
\begin{itemize}[noitemsep, topsep=0pt]
    \item \textbf{GSM8K} \citep{cobbe2021trainingverifierssolvemath}: A benchmark of 8,500 grade-school mathematical word problems requiring multi-step arithmetic reasoning.
    \item \textbf{MATH500} \citep{lightman2023letsverifystepstep}: A subset of 500 challenging problems from the MATH benchmark spanning prealgebra, algebra, geometry, number theory, and calculus.
    \item \textbf{Countdown} \citep{pan2025tinyzero}: A combinatorial reasoning task where models must use three provided numbers and arithmetic operations ($+, -, *, /)$ to construct an expression evaluating to a target number.
    \item \textbf{xLAM} \citep{liu2024apigenautomatedpipelinegenerating}: Function calling dataset consisting of single-tool, multi-tool, and parallel invocation. This dataset was used for SFT and RL training only, not evaluation.
    \item \textbf{BFCL / Tau2} \citep{barres2025tau2benchevaluatingconversationalagents, patil2025the}: A function calling benchmark evaluating single-tool invocation, multi-tool composition, and parallel function, and a complex multi-domain agent evaluation suite consisting of simulated Retail, Airline, and Telecom environments.
\end{itemize}

\paragraph{Baselines.}
We benchmark our framework against representative training configurations.
\begin{enumerate}[noitemsep, topsep=0pt]
    \item \textbf{LLaDA-7B-A1B-Instruct}: The base instruction-tuned diffusion model without additional fine-tuning.
    \item \textbf{+ SFT}: Supervised fine-tuning on teacher reasoning traces generated by Gemini 3.1 Pro.
    \item \textbf{+ RL (diffu-GRPO)} \citep{zhao2025d1scalingreasoningdiffusion}: Standard diffu-GRPO trained from an unconditioned noise canvas $\mathbf{x}_T$.
    \item \textbf{+ RL (Ours)}: CanvasAnneal applied directly to the base model without supervised warm-up.
    \item \textbf{+ SFT \& RL (d1-LLaDA)} \citep{zhao2025d1scalingreasoningdiffusion}: Standard two-stage pipeline consisting of SFT followed by diffu-GRPO.
    \item \textbf{+ SFT \& RL (Ours)}: SFT initialization followed by CanvasAnneal.
\end{enumerate}

\paragraph{Training and Implementation Details.}
Full training configurations, SFT parameters, and multi-component reward functions are detailed in Appendix~\ref{sec:training_details}.

\subsection{Results}

\paragraph{Mathematical Reasoning Performance.}
Table~\ref{tab:math_rl_results} reports mathematical reasoning accuracy across generation lengths. CanvasAnneal improves over standard diffu-GRPO on MATH500 at all three lengths, with gains of $6.0$, $2.0$, and $0.4$ percentage points at lengths 128, 256, and 512, respectively. It also improves Countdown by $2.73$, $3.91$, and $1.96$ points. On GSM8K, however, standard diffu-GRPO remains stronger across all three generation lengths. These results suggest that the effect of curriculum-guided initialization depends on the task, with larger improvements on MATH500 and Countdown than on GSM8K.

CanvasAnneal without SFT also exceeds the SFT+RL d1-LLaDA baseline in several settings. On MATH500, it performs better at lengths 256 and 512, obtaining $41.00\%$ and $45.20\%$ compared with $37.60\%$ and $36.80\%$. On Countdown, it similarly performs better at lengths 256 and 512. However, the SFT+RL baselines remain competitive or stronger in several other configurations, particularly on GSM8K. Thus, teacher-guided RL can reduce the benefit of a separate SFT stage on some tasks, but does not uniformly replace SFT across the evaluated settings.

Combining SFT with CanvasAnneal yields the best result in several individual configurations, including $70.96\%$ on GSM8K at length 256, $36.20\%$ on MATH500 at length 128, and $48.05\%$ on Countdown at length 128. The benefit is again task- and length-dependent rather than uniform.

\begin{table}[t]
\caption{Performance comparison across different math reasoning benchmarks (Base and RL models). We train on the training subset of GSM8K \citep{cobbe2021trainingverifierssolvemath}, MATH \citep{lightman2023letsverifystepstep}, and Countdown \citep{pan2025tinyzero} and evaluate on their evaluation subsets. We perform RL training on a generation length of 256 with denoising steps 128. Evaluation is done at checkpoint 1600 for all RL runs. Reported values are test set accuracy (\%); 95\% bootstrap confidence intervals across test items are approximately $\pm 2.5\%$ for GSM8K, $\pm 4.3\%$ for MATH500, and $\pm 6.1\%$ for Countdown.}
\label{tab:math_rl_results}
\centering
\resizebox{\textwidth}{!}{%
\begin{tabular}{l *{9}{c}}
\toprule
\multirow{2}{*}{\textbf{Model}} & \multicolumn{3}{c}{\textbf{GSM8K}} & \multicolumn{3}{c}{\textbf{MATH500}} & \multicolumn{3}{c}{\textbf{Countdown}} \\
\cmidrule(lr){2-4} \cmidrule(lr){5-7} \cmidrule(lr){8-10}
& \textbf{128} & \textbf{256} & \textbf{512} & \textbf{128} & \textbf{256} & \textbf{512} & \textbf{128} & \textbf{256} & \textbf{512} \\
\midrule
LLaDA-7B-A1B-Instruct & 51.31 &  62.79 & 61.12 & 25.00 & 36.40 & 41.20 &  42.19 &  37.50 & 39.84 \\
    + SFT & 51.86 & 57.47 & 55.19 & 31.60 & 38.40 & 40.40 & 39.45 &  42.97 & 44.92 \\
+ RL (diffu-GRPO)  & 60.65 & 69.98 & 68.23 & 27.80 & 39.00 & 44.80 & 42.58 & 45.70 & 51.95 \\
\textbf{+ RL (CanvasAnneal)} & 59.24 & 68.54 & 66.64 & 33.80 & \textbf{41.00} & \textbf{45.20} & 45.31 & \textbf{49.61} & \textbf{53.91} \\
+ SFT \& RL (d1-LLaDA)  & \textbf{63.46} & 70.51 & \textbf{70.43} & 34.20 & 37.60 & 36.80 & 45.70 & 46.48 & 49.22 \\
\textbf{+ SFT \& RL (CanvasAnneal)}  & 63.23 & \textbf{70.96} & 70.13 & \textbf{36.20} & 37.20 & 41.80 & \textbf{48.05} & 48.44 & 44.14 \\
\bottomrule
\end{tabular}%
}
\end{table}

\paragraph{Tool Use and Function Calling.}
Table~\ref{tab:tool_use_rl_results} reports performance on Tau2 and BFCL. On Tau2, CanvasAnneal consistently outperforms diffu-GRPO across all three domains, improving Retail from $6.14\%$ to $7.90\%$, Airline from $12.00\%$ to $18.00\%$, and Telecom from $12.30\%$ to $13.20\%$. This yields an average score of $13.03\%$, compared with $10.15\%$ for diffu-GRPO and $9.05\%$ for the base model. On BFCL, the results are mixed. CanvasAnneal improves Simple from $53.33\%$ to $59.08\%$ and performs comparably on Multiple, obtaining $65.50\%$ compared with $66.00\%$. However, it underperforms diffu-GRPO on Parallel, with $60.50\%$ compared with $65.00\%$. Overall, these results indicate that CanvasAnneal provides consistent gains on the evaluated conversational tool-use tasks while maintaining competitive, but task-dependent, performance on function calling.

\begin{table}[t]
\caption{Performance comparison on tool-use benchmarks (Tau2 and BFCL) for Base and RL models. We train on xLAM \citep{liu2024apigenautomatedpipelinegenerating} for function calling and evaluate on Tau2 \citep{barres2025tau2benchevaluatingconversationalagents} and BFCL \citep{patil2025the}. BFCL evaluation is done via AST matching. We evaluate on generation length 512.}
\label{tab:tool_use_rl_results}
\centering
\begin{tabular}{l cccc ccc}
\toprule
\multirow{2}{*}{\textbf{Model}} & \multicolumn{4}{c}{\textbf{Tau2}} & \multicolumn{3}{c}{\textbf{BFCL}} \\
\cmidrule(lr){2-5} \cmidrule(lr){6-8}
& \textbf{Retail} & \textbf{Airline} & \textbf{Telecom} & \textbf{Avg} & \textbf{Simple} & \textbf{Multiple} & \textbf{Parallel} \\
\midrule
LLaDA-7B-A1B-Instruct & 5.26 & 14.00 & 7.90 & 9.05 & 56.17 & 63.50 & 64.00 \\
+ SFT & 6.14 & 12.00 & 7.02 & 8.39 & 57.25 & \textbf{77.00} & 50.50 \\
+ RL (diffu-GRPO) & 6.14 & 12.00 & 12.30 & 10.15 & 53.33 & 66.00 & \textbf{65.00} \\
\textbf{+ RL (CanvasAnneal)} & \textbf{7.90} & \textbf{18.00} & \textbf{13.20} & \textbf{13.03} & \textbf{59.08} & 65.50 & 60.50 \\
\bottomrule
\end{tabular}
\end{table}

\begin{figure}[t]
\centering
\includegraphics[width=\linewidth]{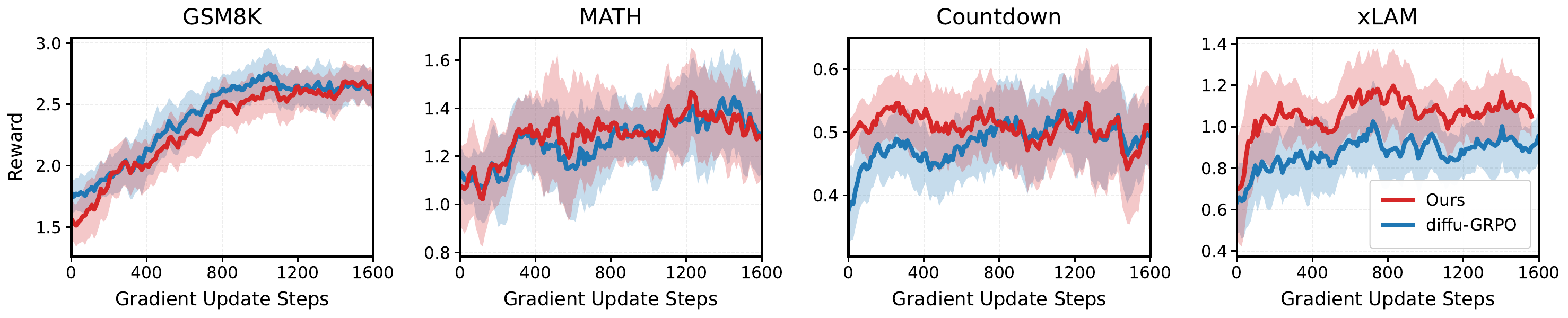}
\caption{\textbf{Training reward curves across gradient-update steps.}
Solid curves show exponential moving averages with weight $0.85$, and shaded regions indicate standard-deviation bounds. CanvasAnneal shows faster reward improvement on xLAM and Countdown. Differences on MATH are smaller, while standard diffu-GRPO performs better on GSM8K.}
\label{fig:reward_curves}
\end{figure}

\paragraph{Training Dynamics.}
Figure~\ref{fig:reward_curves} and Table~\ref{tab:rl_rewards_comparison} compare optimization dynamics over the first 1,600 gradient updates. On xLAM, CanvasAnneal reaches a reward of $1.0$ in $7.4\times$ fewer steps and achieves higher mean and final rewards than standard diffu-GRPO. On Countdown, it reaches a reward of $0.5$ in $18.8\times$ fewer steps and obtains modestly higher mean and final rewards.

On MATH, CanvasAnneal reaches its peak 228 updates earlier and attains a slightly higher peak and mean reward, although its final reward is $0.6\%$ lower. On GSM8K, standard diffu-GRPO obtains higher peak, mean, and final rewards. Taken together, these results suggest that curriculum-guided initialization can improve early optimization and convergence on some tasks, but the effect is not uniform across benchmarks.

\subsection{Ablation Studies}

\paragraph{Prompt-Group Mixture Ratio.}
Table~\ref{tab:mixture_ablation} examines how the fraction of pure-noise prompt groups affects downstream performance. Training with teacher traces for every prompt group performs well in some long-context settings, including MATH500 at length 512, but performs worse in several shorter-generation settings. Increasing the proportion of pure-noise groups generally improves GSM8K and MATH500 performance at shorter lengths.

The 75\% No-Trace / 25\% Injected configuration obtains the strongest GSM8K result at length 128 and the strongest MATH500 result at length 256, while also performing strongly on Countdown at lengths 256 and 512. However, other mixture ratios are preferable in several individual settings. Overall, the ablation suggests that retaining a substantial fraction of pure-noise rollouts is useful when the model is evaluated without teacher context, while some trace-initialized groups can still provide useful training signal.

\begin{table}[t!]
\caption{Ablation of reasoning injection ratio across benchmarks and sequence lengths during GRPO. Reasoning trace injection is controlled at the group level: a ratio of $x$ (e.g., 0.75) indicates $x\%$ of prompt groups receive \textbf{no injected reasoning trace} (pure diffusion generation), while the remaining $(1-x)\%$ receive a partially masked trace where a suffix percentage is uniformly sampled from $\mathcal{U}(0.25, 1.0)$.}
\label{tab:mixture_ablation}
\centering
\begin{tabular}{l *{9}{c}}
\toprule
\multirow{2}{*}{\textbf{Trace Injection Setup}} & \multicolumn{3}{c}{\textbf{GSM8K}} & \multicolumn{3}{c}{\textbf{MATH500}} & \multicolumn{3}{c}{\textbf{Countdown}} \\
\cmidrule(lr){2-4} \cmidrule(lr){5-7} \cmidrule(lr){8-10}
& \textbf{128} & \textbf{256} & \textbf{512} & \textbf{128} & \textbf{256} & \textbf{512} & \textbf{128} & \textbf{256} & \textbf{512} \\
\midrule
0\% No-Trace (100\% Injected) & 52.00 & 64.59 & 65.35 & 27.00 & 38.00 & \textbf{46.20} & \textbf{46.48} & 41.80 & 51.95 \\
25\% No-Trace / 75\% Injected & 57.01 & 66.41 & \textbf{66.72} & 30.40 & 39.80 & 45.20 & \textbf{46.48} & 46.09 & 47.66 \\
50\% No-Trace / 50\% Injected & 57.09 & 65.20 & 63.61 & 29.80 & 39.80 & 43.60 & 43.36 & 38.28 & \textbf{55.08} \\
75\% No-Trace / 25\% Injected & \textbf{60.80} & \textbf{66.41} & 66.19 & \textbf{30.40} & \textbf{40.80} & 45.00 & 41.80 & \textbf{51.95} & 53.91 \\
\bottomrule
\end{tabular}
\end{table}

\paragraph{Curriculum Annealing Schedules.}
Table~\ref{tab:standard_curriculum} compares several masking schedules at a generation length of 256 tokens. The dynamic Beta schedules consistently outperform uniform sampling on MATH, with the $\alpha:1\rightarrow10$, $\beta:10\rightarrow1$ schedule achieving the highest MATH accuracy of $39.00\%$. This schedule also obtains the strongest Countdown result, $44.53\%$, among the evaluated configurations.

The effect is not uniform across tasks. Uniform sampling performs best on GSM8K, reaching $66.34\%$, whereas the dynamic schedules range from $60.42\%$ to $64.52\%$. These results indicate that progressively shifting probability mass toward larger masking ratios can improve performance on some reasoning tasks, but the preferred schedule remains task-dependent.

\begin{table}[t!]
\caption{Ablation of curriculum methods with random masking. We vary the start and end parameters of the Beta distribution to test smooth vs. aggressive transitions over the training steps. Evaluations are conducted with a generation length of 256 tokens.}
\label{tab:standard_curriculum}
\centering
\begin{tabular}{lccc}
\toprule
\textbf{Method} & \textbf{GSM8K} & \textbf{MATH} & \textbf{Countdown} \\
\midrule
Uniform Sampling (0.5, 1.0) & \textbf{66.34} & 34.20 & 43.36 \\
Beta Sampling ($\alpha: 5,\ \beta: 5 \to 1$) & 61.79 & 36.20 & 44.14 \\
Beta Sampling ($\alpha: 1 \to 5,\ \beta: 5 \to 1$) & 60.58 & 37.80 & 42.19 \\
Beta Sampling ($\alpha: 10,\ \beta: 10 \to 1$) & 64.52 & 38.60 & 41.80 \\
Beta Sampling ($\alpha: 1 \to 10,\ \beta: 10 \to 1$) & 60.42 & \textbf{39.00} & \textbf{44.53} \\
\bottomrule
\end{tabular}
\end{table}

\textbf{Suffix Masking versus Random Masking.} Suffix masking preserves a contiguous prefix of the teacher rationale, whereas random masking removes tokens from arbitrary positions. The resulting training states differ structurally. Suffix masking requires the policy to continue a partially observed reasoning trajectory, while random masking additionally requires reconstruction of missing intermediate tokens. We quantify this structural difference by comparing the default dynamic curriculum at a generation length of 256 tokens across both strategies. As shown in Table \ref{tab:math_rl_results}, suffix masking achieves 68.54\%, 41.00\%, and 49.61\% on GSM8K, MATH500, and Countdown, respectively. In contrast, applying the identical curriculum schedule with random masking in Table \ref{tab:standard_curriculum} results in 60.58\%, 37.80\%, and 42.19\% on GSM8K, MATH, and Countdown, respectively. These observations demonstrate that preserving a contiguous reasoning prefix outperforms independent random masking for sequential reasoning tasks, leading to consistent performance improvements.

Our ablations study the effects of trace-injection frequency, curriculum schedule, and masking structure but do not exhaustively evaluate all interactions among these components. In particular, some schedule ablations use random rather than suffix masking. A full factorial decomposition of these design choices is left to future work.

\FloatBarrier

\subsection{Discussion}

Our experiments suggest several properties of curriculum-guided reinforcement learning for discrete diffusion language models.

\begin{itemize}
\item \textbf{Training-Time Guidance without Teacher Inference.}
Teacher traces can be used to provide additional structure during RL training without requiring a teacher model at test time. Because inference begins from a fully masked canvas, the resulting policy retains the standard diffusion generation procedure.

\item \textbf{Optimization Benefits are Task-Dependent.}
Curriculum-guided initialization improves early training dynamics on xLAM and Countdown and provides smaller benefits on MATH. We do not observe the same improvement on GSM8K. Similarly, downstream accuracy improves consistently on MATH500 and Countdown but not on GSM8K or across all tool-use settings. The effectiveness of the curriculum therefore appears to depend on the underlying task and reward landscape.

\item \textbf{Maintaining Pure-Noise Training Examples is Important.}
The mixture ablation indicates that training exclusively from trace-initialized canvases is generally insufficient. Including prompt groups initialized from fully masked canvases exposes the policy to the same type of initialization used at inference and improves performance in several settings. This suggests that teacher-guided rollouts are most useful as a supplement to, rather than a replacement for, unassisted rollouts.

\item \textbf{Structured Masking Provides a Natural Reasoning Curriculum.}
Suffix masking provides a simple mechanism for varying reasoning difficulty while preserving a contiguous prefix of the teacher trajectory. Lower masking ratios require the policy to complete only the final portion of a rationale, whereas larger ratios require increasingly independent generation. Our results indicate that this curriculum can improve optimization and final accuracy on several tasks, although the gains are not universal.

\end{itemize}

\section{Conclusion}

We introduced CanvasAnneal, a reinforcement-learning procedure for discrete diffusion language models that uses partially observed teacher reasoning traces as training-time scaffolds. The method progressively increases the masking ratio of the teacher trajectory while retaining a fixed fraction of fully masked prompt groups throughout training. At inference time, the teacher is removed and generation proceeds from a fully masked canvas using the standard diffusion process.

Across mathematical reasoning benchmarks, CanvasAnneal improves over standard diffu-GRPO on MATH500 and Countdown across the evaluated generation lengths, but does not improve GSM8K. Training-curve analysis similarly shows faster reward improvement on xLAM and Countdown, smaller differences on MATH, and weaker performance on GSM8K. Tool calling performance is consistently improved on Tau2, with competitive performance in BFCL. Overall, these experiments suggest that curriculum-guided initialization can improve diffusion RL optimization and downstream accuracy in some settings, but that its effectiveness depends on the task and training distribution.

\section*{Acknowledgments}

We thank the Gemma 4 Diffusion team for helpful discussions and feedback throughout this work. In particular, we thank Ivan Lobov, João Gante, and Brendan O'Donoghue for their insights and technical feedback on diffusion language models.

\bibliography{main}
\bibliographystyle{tmlr}

\clearpage

\appendix
\section{Training and Implementation Details}
\label{sec:training_details}

All models are trained on 8 $\times$ NVIDIA A100 (80GB) GPUs. Table~\ref{tab:hyperparameters} summarizes the core hyperparameters used for Supervised Fine-Tuning (SFT) and Reinforcement Learning (RL) stages.

\begin{table}[h]
\centering
\caption{Hyperparameters for SFT, RL, and Curriculum schedules.}
\label{tab:hyperparameters}
\resizebox{1.0\textwidth}{!}{%
\begin{tabular}{ll}
\toprule
\textbf{Parameter} & \textbf{Value} \\
\midrule
\multicolumn{2}{c}{\textbf{Supervised Fine-Tuning (SFT)}} \\
\midrule
Global Batch Size & 32 \\
Learning Rate & $1 \times 10^{-5}$ \\
Epochs & 3 \\
Dataset Sizes & GSM8K: 7,470, MATH: 7,497, Countdown: 7,500, xLAM: 7,499 \\
\midrule
\multicolumn{2}{c}{\textbf{Reinforcement Learning (RL)}} \\
\midrule
Optimizer & AdamW ($\beta_1=0.9$, $\beta_2=0.99$, weight decay $0.1$) \\
Learning Rate Schedule & Constant with warmup ratio $0.0001$ \\
Peak Learning Rate & $3 \times 10^{-6}$ \\
Total RL Steps & 1,600 \\
Prompt Batch Size ($B$) & 16 \\
Rollout Batch Size ($B \times G$) & 96 (where $G=6$) \\
PPO Clipping ($\epsilon$) & 0.2 \\
Training Max Generation Length & 256 tokens \\
Training Denoising Steps & 128 \\
\midrule
\multicolumn{2}{c}{\textbf{Curriculum Schedule}} \\
\midrule
Pure Noise Probability ($p_{\text{pure}}$) & $0.25$ for xLAM. $0.75$ for GSM8K, MATH, and Countdown\\
Beta Distribution $\alpha$ & $1.0 \rightarrow 5.0$ \\
Beta Distribution $\beta$ & $5.0 \rightarrow 1.0$ \\
\bottomrule
\end{tabular}
}
\end{table}

At evaluation time, we additionally evaluate generation lengths of 128 and 512 with 64 and 256 denoising steps, respectively. For generation, we employ low-confidence remasking: at each denoising step, we rank the model's predicted tokens by their softmax probability $p_\theta(x_0 \mid \mathbf{x}_t)$ and selectively unmask the highest confidence tokens while keeping the remainder masked.

\subsection{Reward Functions}
To evaluate candidate completions against ground-truth answers, we define multi-component reward functions tailored to each task:
\begin{itemize}
    \item \textbf{GSM8K}: $R = 2.0 \cdot R_{\text{correct}} + 0.5 \cdot R_{\text{strict\_format}} + 0.5 \cdot R_{\text{soft\_format}} + 0.5 \cdot R_{\text{int\_output}} + R_{\text{xml\_count}}$. The $R_{\text{xml\_count}}$ grants $0.125$ for each correctly matched XML tag (e.g., \verb|<reasoning>|, \verb|</reasoning>|).
    \item \textbf{MATH}: $R = 2.0 \cdot R_{\text{correct}} + 0.5 \cdot R_{\text{format}}$, where the format reward requires both \verb|\boxed{}| and answer tags.
    \item \textbf{Countdown}: $R = 1.0 \cdot R_{\text{correct\_eval}} + 0.1 \cdot R_{\text{valid\_format}}$. Formats are verified by validating that the generated arithmetic expression uses the provided numbers exactly once.
    \item \textbf{xLAM}: $R = 2.0 \cdot R_{\text{correct\_json}} + 0.5 \cdot R_{\text{strict\_format}} + R_{\text{xml\_count}}$. The strict format enforces valid \verb|<think>| blocks and proper \verb|<tool_call>| json syntax.
\end{itemize}

\section{Quantitative Analysis of RL Training Dynamics}
\label{sec:rl_dynamics}

To better understand the optimization benefits provided by curriculum-guided initialization, we track the raw reward trajectories over the course of 1,600 PPO updates. Table~\ref{tab:rl_rewards_comparison} provides a quantitative summary comparing our CanvasAnneal procedure against standard diffu-GRPO. 

Our method demonstrates substantial improvements in sample efficiency and peak performance on several tasks. On xLAM, CanvasAnneal achieves a $+16.8\%$ higher peak reward and reaches a reward of 1.0 approximately $7.4\times$ faster than the baseline. Similarly, on Countdown, our method converges to a 0.5 reward threshold $18.8\times$ faster. On MATH, our method reaches its peak performance 228 steps earlier with a marginally higher peak reward. Consistent with our downstream accuracy results, GSM8K does not benefit from the curriculum, exhibiting slightly lower peak and mean rewards during training. These quantitative training dynamics reinforce the observation that reasoning trace initialization provides significant but task-dependent optimization advantages.

\begin{table}[H]
\centering
\caption{Quantitative summary of RL training performance (Steps 1--1,600) comparing \textbf{Ours} against \textbf{diffu-GRPO}.}
\label{tab:rl_rewards_comparison}
\resizebox{\textwidth}{!}{%
\begin{tabular}{l l c c c c}
\toprule
\textbf{Benchmark} & \textbf{Method} & \textbf{Peak Reward} & \textbf{Peak Step} & \textbf{Mean Raw (1--1600)} & \textbf{Final Reward@1600} \\
\midrule
\multirow{3}{*}{\textbf{xLAM}} 
  & diffu-GRPO & 1.0249 & 721 & 0.8883 & 0.9516 \\
  & Ours       & \textbf{1.1975} & 829 & \textbf{1.0762} & \textbf{1.0488} \\
  & \textit{Relative Gain ($\Delta$)} & \textcolor{gainGreen}{\textbf{+16.8\%}} & \textcolor{gainGreen}{\textbf{7.4$\times$ faster to 1.0}} & \textcolor{gainGreen}{\textbf{+21.2\%}} & \textcolor{gainGreen}{\textbf{+10.2\%}} \\
\midrule
\multirow{3}{*}{\textbf{Countdown}} 
  & diffu-GRPO & 0.5371 & 1261 & 0.4903 & 0.4944 \\
  & Ours       & \textbf{0.5468} & 1261 & \textbf{0.5092} & \textbf{0.5043} \\
  & \textit{Relative Gain ($\Delta$)} & \textcolor{gainGreen}{\textbf{+1.8\%}} & \textcolor{gainGreen}{\textbf{18.8$\times$ faster to 0.5}} & \textcolor{gainGreen}{\textbf{+3.8\%}} & \textcolor{gainGreen}{\textbf{+2.0\%}} \\
\midrule
\multirow{3}{*}{\textbf{MATH}} 
  & diffu-GRPO & 1.4452 & 1453 & 1.2697 & \textbf{1.2862} \\
  & Ours       & \textbf{1.4672} & \textbf{1225} & \textbf{1.2935} & 1.2780 \\
  & \textit{Relative Gain ($\Delta$)} & \textcolor{gainGreen}{\textbf{+1.5\%}} & \textcolor{gainGreen}{\textbf{228 steps earlier}} & \textcolor{gainGreen}{\textbf{+1.9\%}} & -0.6\% \\
\midrule
\multirow{3}{*}{\textbf{GSM8K}} 
  & diffu-GRPO & \textbf{2.7533} & 1045 & \textbf{2.4303} & \textbf{2.6063} \\
  & Ours       & 2.6892 & 1549 & 2.3513 & 2.5927 \\
  & \textit{Relative Gain ($\Delta$)} & -2.3\% & -- & -3.3\% & -0.5\% \\
\bottomrule
\end{tabular}
}
\end{table}

\end{document}